\documentclass[10pt,twocolumn,letterpaper]{article}

\usepackage[pagenumbers]{cvpr} 

\usepackage{tabularx}
\usepackage{adjustbox}
\usepackage{booktabs}
\usepackage{multirow}
\usepackage{array}
\usepackage{longtable}
\usepackage{ltxtable}
\usepackage{rotating}
\usepackage{pdflscape}
\usepackage{afterpage}
\usepackage{makecell}
\usepackage{placeins}
\usepackage[accsupp]{axessibility} 

\usepackage[skip=2pt]{caption}

\usepackage{titlesec}
\titleformat{\section}
  {\normalfont\large\bfseries}
  {\thesection}{0.5em}{}

\titleformat{\subsection}
  {\normalfont\normalsize\bfseries}
  {\thesubsection}{0.5em}{}

\titleformat{\subsubsection}
  {\normalfont\normalsize\itshape}
  {\thesubsubsection}{0.5em}{}

\titlespacing*{\section}{0pt}{6pt}{3pt}
\titlespacing*{\subsection}{0pt}{4pt}{2pt}
\titlespacing*{\subsubsection}{0pt}{3pt}{1pt}

\usepackage[font=footnotesize]{caption}
\definecolor{cvprblue}{rgb}{0.21,0.49,0.74}
\usepackage[pagebackref,breaklinks,colorlinks,allcolors=cvprblue]{hyperref}

\def\paperID{13123} 
\def\confName{CVPR}
\def\confYear{2026}

\title{CHIRP dataset: towards long-term, individual-level, behavioral monitoring of bird populations in the wild}

\author{
	Alex Hoi Hang Chan$^{123}$, Neha Singhal$^{4}$, Onur Kocahan$^{4}$, Andrea Meltzer$^{1235}$, Saverio Lubrano$^{1235}$\\ 
	Miyako H. Warrington$^{56}$, Michael Griesser$^{12357*}$, Fumihiro Kano$^{123*}$, Hemal Naik$^{138*}$\\[2ex]
	\normalsize $^1$Centre for the Advanced Study of Collective Behaviour, University of Konstanz \\
	\normalsize $^2$Dept. of Collective Behavior, Max Planck Institute of Animal Behavior \\
	\normalsize $^3$Dept. of Biology, University of Konstanz, 
	\normalsize $^4$Dept. of Computer and Information Science, University of Konstanz \\
	\normalsize $^5$Luondua Boreal Field Station\\
	\normalsize $^6$School of Biological and Medical Sciences, Oxford Brookes University \\
	\normalsize $^7$Dept. of Zoology, Stockholm University \\
	\normalsize $^8$Dept. of Ecology of Animal Societies, Max Planck Institute of Animal Behavior\\[1ex]
	{\small $^{*}$shared senior authorship.}\\ 
	{\small \texttt{\{hoi-hang.chan, andrea.meltzer, saverio.lubrano, michael.griesser, fumihiro.kano\}@uni-konstanz.de}} \\
	{\small \texttt{\{onurkocahan, neha.singhal.blr\}@gmail.com, mwarrington@brookes.ac.uk, hnaik@ab.mpg.de}}
}

\begin{document}
\maketitle
\begin{abstract}
Long-term behavioral monitoring of individual animals is crucial for studying behavioral changes that occurs over different time scales, especially for conservation and evolutionary biology. Computer vision methods have proven to benefit biodiversity monitoring, but automated behavior monitoring in wild populations remains challenging. This stems from the lack of datasets that cover a range of computer vision tasks necessary to extract biologically meaningful measurements of individual animals. Here, we introduce such a dataset (CHIRP) with a new method (CORVID) for individual re-identification of wild birds. The CHIRP (\textbf{C}ombining be\textbf{H}aviour, \textbf{I}ndividual \textbf{R}e-identification and \textbf{P}ostures) dataset is curated from a long-term population of wild Siberian jays studied in Swedish Lapland, supporting re-identification (re-id), action recognition, 2D keypoint estimation, object detection, and instance segmentation. In addition to traditional task-specific benchmarking, we introduce application-specific benchmarking with biologically relevant metrics (feeding rates, co-occurrence rates) to evaluate the performance of models in real-world use cases. Finally, we present \textbf{CORVID} (\textbf{CO}lou\textbf{R}-based \textbf{V}ideo re-\textbf{ID}), a novel pipeline for individual identification of birds based on the segmentation and classification of colored leg rings, a widespread approach for visual identification of individual birds. CORVID offers a probability-based id tracking method by matching the detected combination of color rings with a database. We use application-specific benchmarking to show that CORVID outperforms state of the art re-id methods. We hope this work offers the community a blueprint for curating real-world datasets from ethically approved biological studies to bridge the gap between computer vision research and biological applications.
\end{abstract}
\vspace{-15pt}

\begin{figure*}[!htbp]
\centering
\includegraphics[width=\textwidth]{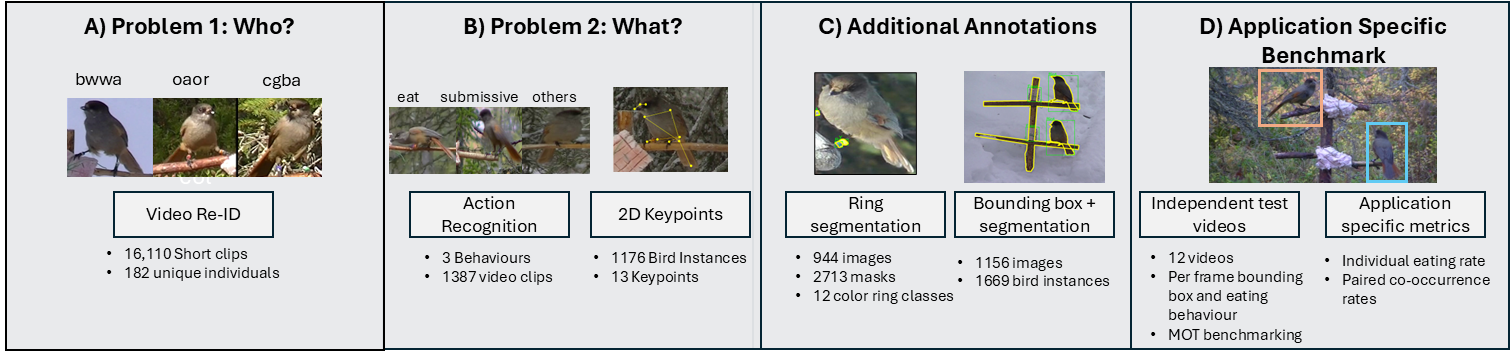}
\caption{\textbf{CHIRP dataset summary.} A) Solving the problem of who, with video re-identification datset of individuals. B) Solving the problem of what, with action recognition dataset and 2D keypoint estimation. C) Additional annotations to support the main tasks, including segmentation (yellow) of color rings, bounding box (green box) and segmentation (yellow) of birds. D) Application specific benchmark, 12 independent test videos with per frame annotation for bounding box, identities and behaviors, with novel metrics on errors related to biological measures like individual feeding rates and paired co-occurrence rates.
}
\label{fig:dataset_overview}
\end{figure*}

\section{Introduction}
Behavior is often the first response of animals when adapting to environmental changes~\cite{tuomainen2011behavioural,sih2011evolution}. Measuring changes in behavior over time is hence crucial for conducting research in behavioral ecology and conservation.
In recent years, rapid developments in technologies have opened new avenues for measuring behavior of wild animals, one of which is the application of computer vision to replace manual observation from images and videos. 

Recent developments in computer vision have shown impressive applications for collecting behavior data in field conditions (outdoor), from 2D \cite{ye2024superanimal,graving2019deepposekit,koger_automated_2023,wiltshire_deepwild_2023,desai_openapepose_2023} and 3D posture estimation \cite{waldmann20233d,chimento2025peering,joska2021acinoset} to action recognition \cite{chan2025yolo,ardoin_automatic_2023,fuchs_asbar_2024,van_de_sande_automated_2024} and individual re-identification \cite{schneider_past_2019-1,kuhl_animal_2013-1,ferreira2020deep}. However, large-scale or long-term deployment of these applications is still limited mainly due to two problems.

Firstly, computer vision research focuses on solving specific tasks such as object detection, re-id or keypoint estimation, whereas long-term monitoring often requires multiple computer vision tasks to be solved simultaneously. In long-term monitoring, biologists are generally interested in identifying \textit{who} did \textit{what} behavior. To achieve automated monitoring, re-identification and behavioral recognition tasks have to be solved together, either directly or using supporting sub-tasks like detection, tracking and keypoint estimation. Therefore, the classic strategy of developing task-specific methods is not suitable for behavioral monitoring because often it is difficult to readily combine and deploy novel algorithms directly for data collection. This challenge can be alleviated by curating large datasets with annotations to solve multiple tasks using the same real-world data. 

Secondly, traditional benchmarking processes lack additional metrics to support deployment for biological applications. While some recently published datasets provide annotations for solving a larger range of computer vision tasks (See supplementary; \cite{3D-POP,ma_chimpact_2023,duporge2024baboonland,ng2022animal,naik2024bucktales,liu2023lote}), benchmarking is often done for each individual task in isolation. This creates ambiguity for users while selecting methods for deployment, because task specific benchmarking does not reflect the impact of choosing a specific method on the final biological measurements. The unpredictable nature of error propagation is demonstrated with recent case studies \cite{pantazis2024deep,chan_towards_2025}, and a possible solution is to include a benchmarking mechanism that allows tasks to be combined and tested.


In this paper, we address these problems and introduce computer vision solutions for long-term, individual-level behavioral monitoring in the wild. Firstly, we present the CHIRP dataset (\textbf{C}ombining be\textbf{H}aviour \textbf{I}ndividual \textbf{R}e-identification and \textbf{P}ostures), the first dataset focusing on social behaviors in a wild bird population, the Siberian jay (\textit{Perisoreus infaustus}), a group living corvid. Secondly, we present the application-specific benchmark, a novel evaluation paradigm that makes use of application-specific metrics like feeding rate and co-occurrence rates of individuals. Our benchmarking allows novel methods to be directly tested for their impact on the final biological measurement, encouraging method optimizations for downstream applications instead of individual tasks. Thirdly, we introduce CORVID (\textbf{CO}lou\textbf{R}-based \textbf{V}ideo re-\textbf{ID}entification), a novel automatic re-id framework as a baseline for the application specific benchmark, by identifying birds via individual color rings. While color leg rings are widely used to mark individuals in wild bird populations \cite{anderson_value_2009,calvo1992review}, to the best of our knowledge, we present the first computer vision framework to automatically identify individual birds using color rings. We hope this work can illustrate how datasets and evaluation procedures can be designed with clear deployment objectives, to allow novel computer vision methods to be better applied for downstream biological understanding.

\begin{figure*}[!htbp]
\centering
\includegraphics[width=0.8\textwidth]{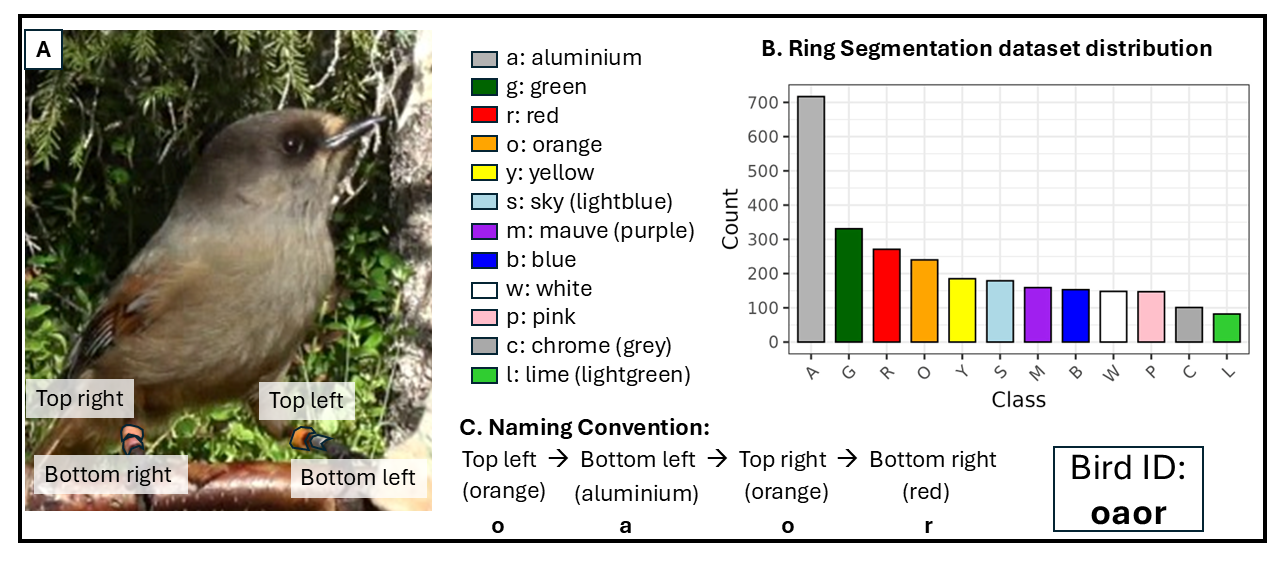}
\caption{\textbf{Summary of leg color ring definitions and naming convention of the Siberian jays.} A) Sample image of an  individual, and a list of the different color rings . Ring positions are defined as top/bottom left/ right from the bird’s perspective. B) Class distribution of ring masks provided in the ring segmentation dataset. C) Naming convention, each bird is named in the order of top left, bottom left, top right, bottom right ring. The color combination of bird in the picture is oaor: orange (o), aluminium (a), orange (o),  red (r). 
}
\label{fig:ringdef}
\end{figure*}


\section{Related Works}
Over the last decade, the popularity of animal-specific datasets has increased significantly. Initial datasets included task-specific data to tackle computer vision problems such as detection and tracking~\cite{zhang_animaltrack_2023,wang_watb_2023}, 2D and 3D keypoint estimation ~\cite{joska2021acinoset,desai_openapepose_2023,marshall2021pair,yu2021ap,li_poses_2024}, action recognition~\cite{brookes_panaf-fgbg_2025,Brookes2024pan,chen2023mammalnet,gabeff_mammalps_2025,kholiavchenko_kabr_2024} and re-identification (see \cite{cermak_wildlifedatasets_2024}). These datasets have contributed to bringing computer vision innovations to animal studies, allow biodiversity monitoring to be scaled and automated \cite{tuia2022perspectives}. However, it is widely acknowledged that long-term studies involve solving multiple computer vision tasks together. In response, recent datasets are moving towards more holistic and real-world datasets that contain multiple computer vision tasks within the same study system (see supplementary).

Datasets such as Animal Kingdom~\cite{ng2022animal} primarily focus on the problem of animal behavior recognition of 850 animal species, with ground truth labels on action recognition tasks, but also for object detection and keypoint estimation. However, the dataset was collated from internet-sourced youtube videos and direct applicability of methods developed with the dataset for biological studies remain unvalidated.

Datasets like LoTE-animal~\cite{liu2023lote} or Baboonland~\cite{duporge2024baboonland} use data directly sourced from biological studies and focuses on action recognition tasks, with support of a wide range of computer vision tasks but do not provide re-id. Similarly, datasets like 3D-POP~\cite{3D-POP} and Bucktales~\cite{naik2024bucktales},  focus on tracking large groups for collective behavior and provide ground truth for bounding box, re-id and 2D/3D postures but no annotations for individual activity. These datasets allow novel algorithms to be better bridged to downstream biological applications because they are curated from the data collected for biological studies. In the CHIRP dataset, we follow the same blueprint to ensure that both behavior and identity annotations are supported for downstream applications.

The WILD dataset~\cite{xiao2023multi} is a dataset from a social behavior study and offers multi-view 3D tracking and re-id of cowbirds in semi-wild environment. While not focusing on action recognition, that dataset proposes bird re-identification, with the bird subjects also fitted with color leg rings. However, the authors did not explicitly make use of color ring information for re-id, instead use an image classifier on an image crop of the birds. Similarly, IndividualBirdID~\cite{ferreira2020deep} also use cropped image of the back patterns of birds for re-id, showing that CNNs can reliably identify individuals across 3 bird species. Here, we include bird re-id problem from an additional perspective of  using color information from the leg rings with CORVID.

Finally, ChimpACT \cite{ma_chimpact_2023} is a dataset on semi-captive chimpanzees in Leipzig Zoo, with ground-truth labels on identities, bounding boxes, postures and behaviors, focusing on longtitudinal (long-term) monitoring of chimpanzee groups. ChimpACT provides task specific benchmarking results~\cite{ma_alphachimp_2024,ma_chimpact_2023}, which makes it hard to evaluate whether individual model performances are sufficient to achieve accurate longtitudinal individual-level behavioral monitoring. CHIRP dataset includes application specific benchmarks, with new metrics designed to evaluate the performance of algorithms directly in the context of the final use case.

\section{The CHIRP Dataset}

To facilitate long-term, individual-level, behavioral monitoring, it is critical to record \textit{who} does \textit{what} (\autoref{fig:dataset_overview}). 
Thus, we first provide a large-scale video re-identification dataset to determine \textit{who} is present in the video frame. Then, to classify \textit{what} the individuals are doing, we provide annotations for action recognition of behavioral classes, and 2D keypoints for fine-scaled kinematics. Below, we first describe the study system and data collection, then detailed descriptions of the provided annotations, and finally the application-specific benchmark that complements the dataset. All data samples in the dataset will be provided with a date of data acquisition, to allow for time-aware splits. We also note that except for the video re-id dataset, all datasets adopt an 80/20 train/test split. 
The complete dataset, code and metadata is available here: \url{https://github.com/alexhang212/CHIRP_Dataset}

\subsection{Data collection}
CHIRP dataset is collected in a long-term study population of Siberian jays (\textit{Perisoreus infaustus}), in Swedish Lapland (65°40' N, 19°0' E) between 2014 and 2022. Siberian jays are group-living birds (group size range: 2-7); groups typically consist of a breeding pair and non-breeder, which are either retained offspring or unrelated immigrants \cite{dickinson_siberian_2016}. Groups are year-round territorial and can be found in predictable locations, allowing reliable monitoring of groups.  Each bird is fitted with an aluminum ring and a unique combination of 2 to 3 colored plastic rings of 11 unique colors (permit via Umeå Animal Ethics Board, A23-20 and A26-13 under the license of the Swedish Museum of Natural History). This provides up to 1331 (11$^3$) possible unique combinations, allowing researchers to visually identify individuals in the field (\autoref{fig:ringdef}A). Color rings ensure monitoring accuracy across lifetime and is not affected by changing plumage of birds. Due to the unique social structure of Siberian jays, the long-term study strives to answer questions related to the evolution of cooperative behavior and social buffering effects towards environmental changes~\cite{griesser2017family,warrington2024stronger}.

In this study, videos are taken as part of a standardized behavioral recording protocol, where 15-30 min long videos (25fps, 1920 x 1080) are recorded in each group using a standardized feeding perch. Then, researchers manually annotate identity and behavior from videos using the BORIS annotation software \cite{friard_boris_2016} e.g. feeding bouts, submissive behaviors, and displacements for each individual (see supplementary for complete ethogram). BORIS~\cite{friard_boris_2016} is a widely used annotation software in animal behavior studies to manually code time of behavior bouts, and these annotations were used to reduce annotation effort. Overall, 443 unique videos (across 9 years, 2014-2022) were used to prepare the CHIRP dataset, and date for each data sample is always provided for time-aware splits. We also ensured that samples from the same behavioral videos are never assigned in the same split for all subsequent datasets. In addition, we also collected multi-view data on jays foraging on the ground, which only contributes to the 2D keypoint dataset.


\subsection{Problem 1: Who is present?}
\subsubsection{Video re-identification dataset}
The video re-identification (re-id) dataset consists of 16,190 video clips (25 frames, 1 second each) for 183 unique individuals (\autoref{fig:dataset_overview}). Each individual has on average 89 samples, and 32.42\% of birds (N= 59) has samples across multiple years. We first train a YOLOv8 object detection model~\cite{YOLO} (see \autoref{sec:add_annot}) to detect birds within each 15-30 minute long behavioral video. Then using manual annotation from BORIS, 1-second long cropped video clips (25 frames) are extracted, in video segments where only 1 bird was present for both BORIS and YOLO, allowing automatic matching of identities. We manually check the video clips to ensure the automatic ID assignment is correct, filtering out 8.3\% false positive video clips due to manual mis-labelling and false positive detections in YOLO. We ensure all short clips from the same 15-30 minute behavioral video are always in separate data splits, to assure no data leakage from background information is present. 

Using the clips of individuals, we prepared a closed split, disjointed split and open split based on definitions formalized by Wildlife Datasets \cite{cermak_wildlifedatasets_2024}. 
The closed set split is closest to the final use-case for the Siberian jay system, as researchers are always present when recording new videos allowing new individuals to be acknowledged and added to the gallery. In the closed-set split, all individuals are both in the train and test set (with 80/20 ratio), and the task is to assign individuals in the test set to individuals in the training set. 
Next, the disjointed split is created to test for generalization across systems. We assign approximately half the individuals to the train and test set respectively. Within the test set, we further assign all clips from a single behavioral video as the gallery, and the rest as query. The re-identification task here is to match query clips to an individual within the gallery database.

Finally, the open-set split is prepared to test for the robustness against new individuals, for alternative deployments that may not involve researchers on site (e.g. passive camera traps). We assign 20\% and 80\% of individuals as "unknown" and "known" respectively. Within the known individuals, data samples are split into train and test set (80/20 ratio), with data samples of the unknown individuals added to the test set. The re-identification task is to determine if a data sample is "known" or "unknown", then if known, assign the correct ID.

Siberian jays are territorial and group members remain fairly consistent within each location, with occasional encounters with neighbouring territories~\cite{griesser_fine-scale_2015}. Taking advantage of this biological feature, we offer two levels of metadata for each video. First, a short-list of individuals within the territory (N=2-4, mean 2.92), and second, a list of individuals that are also likely likely to be seen in that territory (neighbours; N=9-25, mean 14.58). This adds a layer of complexity and varying difficulty to the dataset, allowing domain-specific knowledge to be built into potential computer vision solutions for the re-id problem.


\subsection{Problem 2: What are they doing?}
\subsubsection{Action recognition dataset}
\begin{figure}[!htbp]
\centering
\includegraphics[width=0.7\columnwidth]{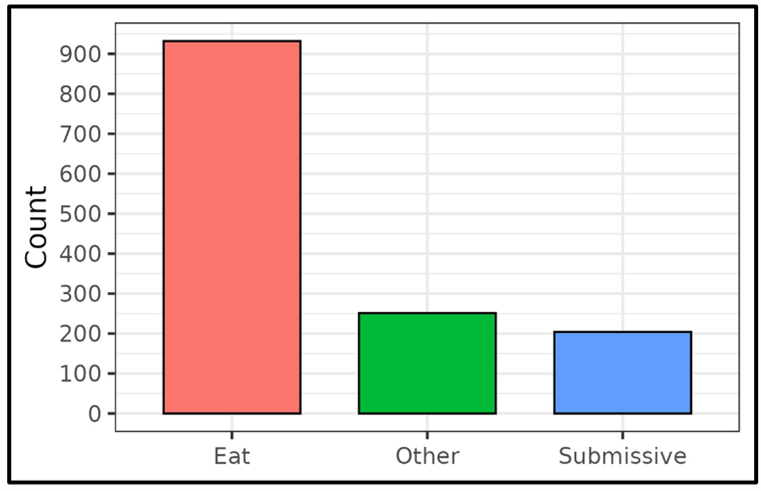}
\caption{\textbf{Class distribution for  action recognition dataset}}
\label{fig:action_rec}
\end{figure}

The action recognition dataset includes manual annotations of 1387 short video crops of birds (min 25 frames, max 74 frames, mean 66.73 frames) performing each behavior (\autoref{fig:dataset_overview}). We code “eat” as a bird that was pecking at the food, “submissive” as individuals displaying stereotyped wing-flapping behavior \cite{griesser_fine-scale_2015}, and “others” as any other behavior, which includes vigilance, resting and flying (see \autoref{fig:action_rec} for class distribution). To speed up annotation, we first train a YOLOv8 model \cite{YOLO} (see \autoref{sec:add_annot}) to extract bird tracks from each video, then use BORIS annotations to identify and extract video segments that contains behaviors of interest. Finally, we manually review and annotate each short clip with the appropriate behavior.

\subsubsection{2D Keypoint estimation dataset}
We provide a comprehensive 2D keypoint dataset of 1176 individual bird instances, across 879 images (\autoref{fig:dataset_overview}), with manually annotated keypoint ground truth for 13 unique keypoints. 36\% of frames are taken from standardized feeding videos, while 64\% are taken from videos from another feeding context, where birds are foraging on the ground. Each instance also includes corresponding bounding box annotation for each individual bird. 

\subsection{Additional annotations}
\label{sec:add_annot}
In addition to the main tasks described above, we also provide additional annotations to support the main tasks. Firstly, we provide annotations for object detection and instance segmentation. For this, we provide 1156 frames of 1669 bird instances with annotated bounding box and segmentation mask of birds and the feeding stick, as well as bounding box around the food. To reduce manual annotation time, we use SAM2 \cite{ravi_sam_2024} to automatically generate masks of birds based on bounding box prompts, which we validate against 688 annotations, with a mean 0.84 intersection over union (IOU). 
Secondly, as the color rings are the most visually salient features that directly encodes individual identities within an image in Siberian jays, we also provide bounding box and segmentation masks of individual color rings. We provide annotations across 944 frames of cropped images of birds, with 2713 unique ring instances of 12 unique color classes (\autoref{fig:ringdef}B).

Finally, to allow for training end-to-end or multi-modal methods, we also provide model-annotated datasets by using current best model (see Section \ref{sec:benchmarking}) to produce 2D keypoints and segmentation for the video re-id and action recognition datasets.

\subsection{Application specific benchmark}
We provide 12 independent test videos of 35 bird individuals with frame-by-frame annotations of eating behavior, identities and bounding boxes for the application specific benchmark. These videos are not present in any other annotations provided, thus no data leakage is possible. We first use YOLOv8 trained on the bounding box annotations above to obtain 2D tracks, then manually assigned bird ID to each track. We match these tracks to additional BORIS annotations, where an observer marks every instance where the beak of a bird individual touches the food when feeding. Since the tracks obtained from YOLO contains segmented tracks (e.g., when an individual jumps from one side to the other), we use linear interpolation to join two tracks that were marked as the same individual. The resulting dataset consists of frame-by-frame bounding box with coded IDs and all instances of feeding by each individual. In addition to acting as an application specific benchmark, this dataset also acts as a multi-object tracking (MOT) benchmark dataset due to the availability of frame-by-frame bounding box annotations with identities.

Since we design the application specific benchmark specifically for the CHIRP dataset, we also introduce novel metrics to evaluate and compare the performance of models for application relevant use cases (see supplementary methods for detailed descriptions). First, we evaluate on two lower-level metrics, 1) proportion correct frame assignments, defined as the proportion of ground truth tracks and frames that are assigned to the correct individual, to evaluate tracking and individual identification performance. 2) We compute mean precision, recall and F1-score for individual feeding events by splitting each video into 1s time windows, with true positives defined as pecking of the given individual detected within the same time window, averaged across all individuals. Next, we also calculate two higher-level biological measures, 1) individual level feeding rates (pecks/minute) and 2) co-occurrence rates, as the proportion of time spent together between each pair of individuals, divided by total video length. For both biological measures, we computed the mean, median and standard deviation of absolute errors and Pearson’s correlation between the predicted and ground truth.

\begin{figure*}[!htbp]
\centering
\includegraphics[width=0.9\textwidth]{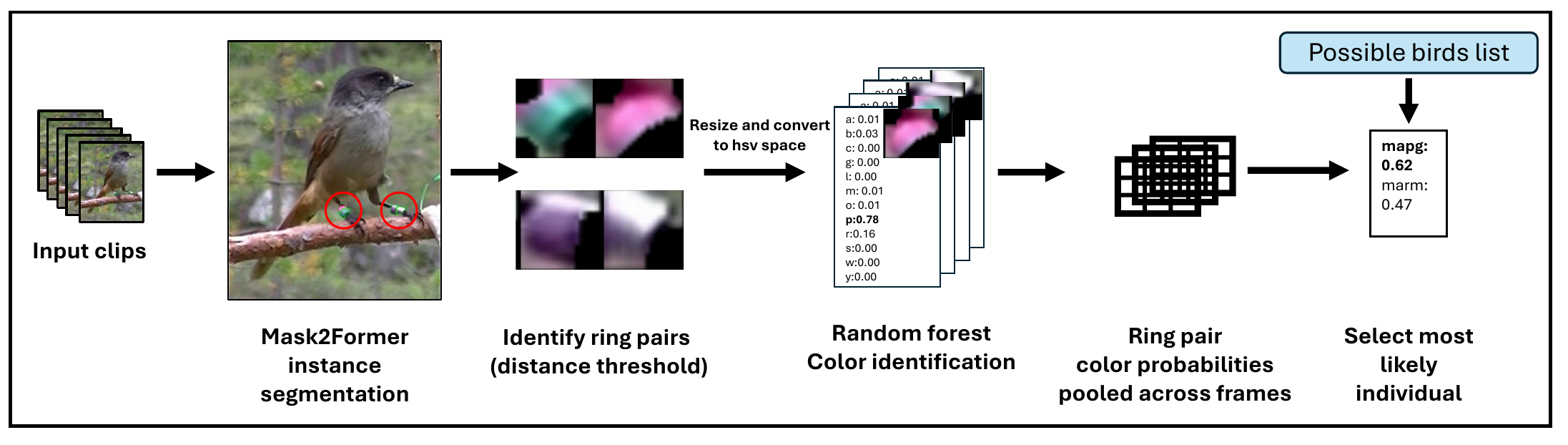}
\caption{\textbf{CORVID pipeline.} Schematic for the color based re-ID approach pipeline. 1 second clips from CHIRP are fed into Mask2Former instance segmentation model, to extract masks of rings. The rings are grouped into ring pairs based on a distance threshold, then resized and converted into hsv space. The images are fed into a random forest model to predict probabilities of each color, then combined with associated ring pair to create a probability matrix of every color pair, then pooled across frames. Finally, the most probable bird is selected based on the possible birds that could be present in a given video.}
\label{fig:COBRA}
\end{figure*}

\section{CORVID: \textbf{CO}lou\textbf{R}-based \textbf{V}ideo re-\textbf{ID}}
To set an initial benchmark for the re-identification task, we propose a novel pipeline that aims to detect individual color rings for individual identification. Attaching unique combinations of color rings to wild birds is common practice for population monitoring (861 "combination of uncoded color leg rings" projects covered by cr-birds database~\cite{cr-birding}) and long-term demographic studies (105/175 populations color ringed covered by SPI-bird database~\cite{culina2021connecting}), thus a method for individual identification based on color rings is widely applicable.
However, to the best of our knowledge, we are the first to directly leverage this feature of bird study systems to achieve individual identification through the detection of color ring patterns. Previous work uses computer vision methods to automatically detect tags that are placed on animals, including color barcodes \cite{nagy2013context,nagy_long-term_2023,santiago-plaza_identification_2024,meyers2023towards}, and fiducial markers (e.g QR codes or aruco tags; \cite{alarcon2018automated,crall_beetag_2015,wolf_naps_2023}. Other work has also explored deep learning based classifiers to recognize bird individuals, both in captivity and in the wild~\cite{ferreira2020deep,xiao2023multi}. However, compared to existing re-id approaches, our approach do not rely on any of the training data provided in the video re-id dataset, and only relies on the detection of color rings. This allows for the method to be the generalized to any new individuals given the ring combinations are known, and no new color is introduced.

The pipeline has three main components (\autoref{fig:COBRA}). Firstly, we detect individual rings using a Mask2Former instance segmentation model~\cite{cheng_masked-attention_2022} trained on the ring segmentation dataset, then cropped, transformed into HSV space, and resized into a 20x20 resolution images. In the second step, we feed color histograms from the images into a random forest model trained on the ring segmentation dataset, to output confidence scores for each color. We formalize the problem as a multi-class classification problem to allow for the model to predict confidence scores for each color class, considering some color classes are similar to each other. As the final step, we implement a matching algorithm by first identifying ring pairs based on centroid distance threshold of each ring detection, then sum up the probability of every paired ring color combination based on the outputs of the random forest classifier, across the 25 video frames. We then match the final score with the possible ID metadata for the data sample, and the most likely ID for a given video clip is identified. We refer to the supplementary section for more details on the pipeline and exploration of the CHIRP ring segmentation dataset.

\section{Benchmarking}
\label{sec:benchmarking}
To explore the performance of state-of-the-art models on the CHIRP dataset, we provide task-specific benchmarks for each of the main tasks proposed. 
Next, we implement a simple pipeline to be applied on the application-specific benchmark, to provide a baseline on how the best algorithms performs when combined for automated data extraction.

\subsection{Video re-identification}
For video re-id, we compare our proposed CORVID pipeline to Mega Descriptor, a foundation model for animal re-identification \cite{cermak_wildlifedatasets_2024}. We compare CORVID with Mega-descriptor pre-trained on other animal datasets, and Mega Descriptor that is fine-tuned with CHIRP (\autoref{tab:Reid_bench}). For Mega Descriptor, we pooled frame-wise probabilities within a tracklet, and the ID with the highest average score was assigned. For benchmarking, we only benchmarked the closed and disjointed split, as we do not have a reliable way of distinguishing between known and unknown individuals using our proposed CORVID method. For each data split, we test three conditions, we use possible birds for the given data sample in the gallery (“within territory”), possible birds with neighbours ("within territory + neighbours) and all birds in the re-id dataset ("All"). We found that CORVID outperforms MegaDescriptor with the within territory and neighbours constraint, but the contrary is true in the closed-set when the gallery include all individuals (\autoref{tab:Reid_bench}). This shows that explicitly detecting and using ring color information from the CORVID pipeline seem to be better than state-of-the-art deep metric learning techniques, but this relies on the within-territory constraint.

\begin{table*}[ht]
\centering
\caption{\textbf{Video Re-ID benchmarks.} We compare CORVID, pre-trained and fine-tuned mega descriptor across different evaluation settings. Bold denotes best performing model for each metric.}
\footnotesize
\renewcommand{\arraystretch}{1.1}
\begin{adjustbox}{}
\begin{tabular}{l@{\hskip 0.3cm}ccc@{\hskip 0.3cm}ccc@{\hskip 0.3cm}ccc@{\hskip 0.3cm}ccc}
\toprule
\multirow{3}{*}{\textbf{Method}}
& \multicolumn{6}{c}{\textbf{Closed set}}
& \multicolumn{6}{c}{\textbf{Disjointed set}} \\
\cmidrule(lr){2-7} \cmidrule(lr){8-13}
& \multicolumn{2}{c}{\makecell{Within\\Territory}}
& \multicolumn{2}{c}{\makecell{Within Terr.\\+ Neighbours}}
& \multicolumn{2}{c}{\textbf{All}}
& \multicolumn{2}{c}{\makecell{Within\\Territory}}
& \multicolumn{2}{c}{\makecell{Within Terr.\\+ Neighbours}}
& \multicolumn{2}{c}{\textbf{All}} \\
\cmidrule(lr){2-3} \cmidrule(lr){4-5} \cmidrule(lr){6-7}
\cmidrule(lr){8-9} \cmidrule(lr){10-11} \cmidrule(lr){12-13}
& Top-1 & Top-3 & Top-1 & Top-3 & Top-1 & Top-3 & Top-1 & Top-3 & Top-1 & Top-3 & Top-1 & Top-3 \\
\midrule
CORVID 
& \textbf{0.66} & \textbf{0.96} 
& \textbf{0.29} & \textbf{0.49 }
& 0.05 & 0.07 
& \textbf{0.69} & \textbf{0.97} 
& \textbf{0.31} & \textbf{0.53}        
& \textbf{0.06} & \textbf{0.13} \\
Pre-trained Mega 
& 0.28 & 0.67
& 0.19 & 0.41
& \textbf{0.10} & \textbf{0.19}
& 0.31 & 0.62 
& 0.14 & 0.32 
& 0.05 & 0.10 \\
Fine-tuned Mega 
& 0.27 & 0.56 
& 0.17 & 0.35 
&\textbf{ 0.10} & 0.17 
& 0.41 & 0.71 
& 0.13 & 0.27 
& 0.05 & 0.09 \\
\bottomrule
\end{tabular}
\end{adjustbox}
\label{tab:Reid_bench}
\end{table*}

\subsection{Action Recognition}
Next, we train a series of models for video action recognition using the MMAction2 library \cite{mmaction2_contributors_openmmlabs_2020}. We benchmark video-based methods here, but posture based methods can also be used. We train each model for 100 epochs, and the best epoch in terms of top 1 accuracy in the test set was chosen (\autoref{tab:actionrec_bench}). Overall, C3D performs the best, with an accuracy of 0.72. 

\begin{table}[ht]
\caption{\textbf{ Action recognition benchmarks.} We compute precision, recall, f1-score and top-1 accuracy for each model. Bold denotes best performing model for each metric.}
\centering
\begin{tabular}{lcccc}
\toprule
\textbf{Model} & \textbf{Precision} & \textbf{Recall} & \textbf{F1} & \textbf{Accuracy} \\
\midrule
SlowFast & 0.460 & 0.678 & 0.548 & 0.678 \\
C3D      & \textbf{0.675} & \textbf{0.715} &\textbf{ 0.684} & \textbf{0.715} \\
X3D      & 0.460 & 0.678 & 0.548 & 0.678 \\
\bottomrule
\end{tabular}
\label{tab:actionrec_bench}
\end{table}

\subsection{2D Keypoint Estimation}
For keypoint estimation, we train models using the MMPose library \cite{mmpose_contributors_openmmlab_2020}. We compute mean and median Euclidean error, root mean squared error (RMSE) and percentage correct keypoints (PCK05, PCK10), as a keypoint estimate that lies within 5\% and 10\% of the largest dimension of the ground truth bounding box. We train all models for 100 epochs, and the best epoch based on test PCK is selected. \autoref{tab:keyopint_bench} shows benchmarking results, with ViTPose large being the best performing model, and all architectures performing well as evident from the high PCK values. This is indicative that the keypoint annotations provided will allow for training of accurate keypoint estimation models for further tasks like action recognition.
\begin{table}[ht]
\caption{\textbf{2D Keypoint estimation benchmarks}. Mean, median and root mean squared error (RMSE) was computed using Euclidean distances between predicted and ground truth keypoints. PCK10 and PCK05 stands for percentage correct keypoints within 10\% and 5\% of ground truth, scaled by the largest dimension of the bounding box. Bold denotes best performing model for each metric.}
\begin{adjustbox}{width=\columnwidth}
\begin{tabular}{lccccc}
\toprule
\textbf{Model} & \textbf{Mean Error (px)} & \textbf{Median Error (px)} & \textbf{RMSE (px)} & \textbf{PCK@10} & \textbf{PCK@5} \\
\midrule
ResNet-50        & 12.03 & 7.568 & 20.65 & 0.940 & 0.832 \\
ResNet-101       & 12.47 & 7.491 & 21.24 & 0.940 & 0.830 \\
ResNet-152       & 10.91 & 7.245 & 17.94 & 0.961 & 0.848 \\
HRNet            & 10.86 & 6.463 & 19.42 & 0.951 & 0.862 \\
ViTPose-small    & 10.32 & 6.872 & 16.70 & 0.957 & 0.859 \\
ViTPose-large    & \textbf{7.773} & \textbf{5.194} & \textbf{12.64} & \textbf{0.978} & \textbf{0.915} \\
\bottomrule
\end{tabular}
\end{adjustbox}
\label{tab:keyopint_bench}
\end{table}
\subsection{Application-specific benchmark}
We implement a simple pipeline that combines different models and algorithms with the aim of extracting individual co-occurrence and feeding rates. The pipeline includes 1) object detection to identify bounding boxes of birds, 2) tracking algorithm to combine detections into tracklets, 3) individual identification for each tracklet and 4) action recognition within each tracklet. As an initial experiment to determine how accuracies in task-specific metrics propagates to application-specific metrics, we compared 3 methods for ID assignment: CORVID, MegaDescriptor fine-tuned on the disjointed set and random assignment (repeated 100 times, to obtain mean estimates). For all three methods, we use a list of possible birds within the territory to constraint the possible birds to a list of 2-5 individuals. As solutions improve in the future, the possible bird list can be expanded to include neighbours, and eventually the whole population. For the rest of the pipeline, we use YOLOv8 for object detection and BoTSORT (using boxmot library; \cite{brostrom_boxmot_nodate}) for tracking. For action recognition, we use the best performing C3D model for action recognition by splitting tracklets into 1s segments.
In addition, we also provide a human benchmark, by coding the first 5-minutes of each video by an independent human coder to provide baseline value, as the target for future models to reach.

\begin{figure*}[!htbp]
\centering
\includegraphics[width=\textwidth]{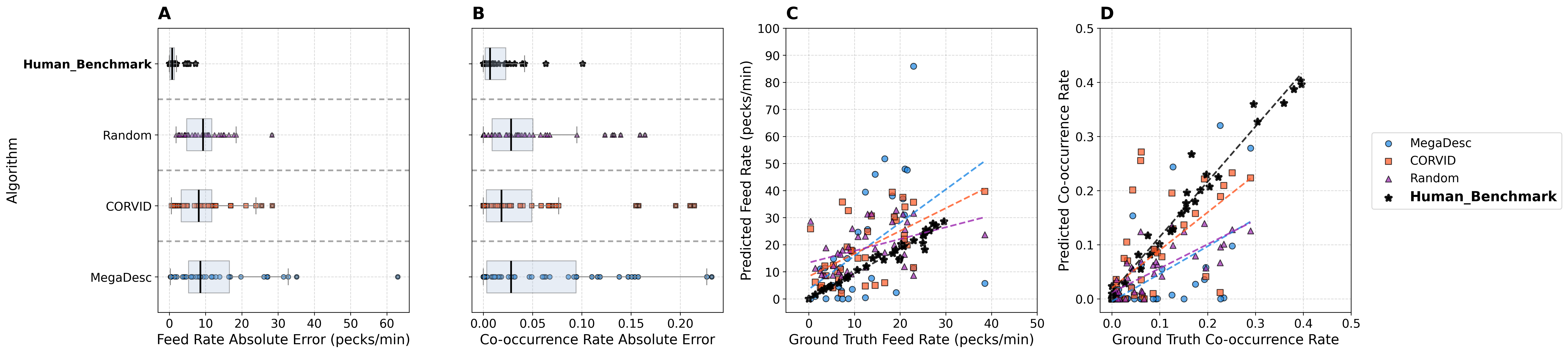}
\caption{\textbf{Application specific benchmark results.} Comparing ground truth measurements and predictions from pipelines to test for how different components affects biological measurements. We compared proposed CORVID pipeline, fine-tuned MegaDescriptor and random assignments for individual recognition. All pipelines used YOLOv8 for object detection, BoTSORT for tracking, and C3D for action recognition. Absolute errors of A) individual-level feeding rates and, B) co-occurrence rates and correlation of C) individual-level feeding rates and, D) co-occurrence rates. Individual feeding rates defined as number of times individual pecks at the food (pecks/min), and co-occurence rates is defined by the proportion of time two individuals were detected together, scaled by video length.}
\label{fig:AppSpec_bench}
\end{figure*}

We find that differences in performance between CORVID and Mega Descriptor in task-specific metrics (\autoref{tab:Reid_bench}) predicts the accuracy in application-specific metrics, as evident from higher performance when using CORVID in the pipeline (\autoref{tab:appspec_lower}, \ref{tab:appspec_upper}). Surprisingly, random assignment performs the best in some metrics (\autoref{tab:appspec_upper}), showing that the proposed methods still have room for improvement. We also find that MegaDescriptor performs worse than random assignment across all higher-level metrics (\autoref{fig:AppSpec_bench}), highlighting that the model is not suitable for deployment. 
Finally, errors from all pipelines are still high compared with the human benchmark in both individual feeding rates and co-occurrence rates (\autoref{fig:AppSpec_bench}, \autoref{tab:appspec_upper}), and further improvements will be needed, highlighting the value of the CHIRP dataset.

\begin{table}[ht]
\centering
\caption{\textbf{Lower-level baseline metrics for application specific benchmark.} Metrics evaluate tracking, ID assignment accuracy, and behavioral recognition performance. Bold denotes best performing model for each metric.}
\renewcommand{\arraystretch}{1.0}
\begin{adjustbox}{width=\columnwidth}
\begin{tabular}{lcccc}
\toprule
\makecell{\textbf{Individual}\\\textbf{Identification}} &
\makecell{\textbf{Prop. correct}\\\textbf{frames}} &
\makecell{\textbf{Peck}\\\textbf{Precision}} &
\makecell{\textbf{Peck}\\\textbf{Recall}} &
\makecell{\textbf{Peck}\\\textbf{F1}} \\
\midrule
CORVID & \textbf{0.647} & \textbf{0.485} & \textbf{0.725} &\textbf{0.537} \\
MegaDescriptor & 0.617 & 0.410 & 0.550 & 0.408 \\
Random  & 0.331 & 0.303 & 0.436 & 0.327 \\

\bottomrule
\end{tabular}
\end{adjustbox}
\label{tab:appspec_lower}
\end{table}

\begin{table}[ht]
\centering
\caption{\textbf{Higher level biologically relevant metrics.} Comparison of individual feeding rates and co-occurrence rates across different pipeline configurations. Bold denotes best performing method for each metric, and arrows represents whether a higher or lower value is better}
\tiny
\renewcommand{\arraystretch}{0.9}
\begin{adjustbox}{width=\columnwidth,center}

\begin{tabular}{l@{\hskip 0.2cm}cccc@{\hskip 0.2cm}cccc}
\toprule
\multirow{2}{*}{\makecell{\textbf{Individual}\\\textbf{ID}}}
& \multicolumn{4}{c}{\textbf{Individual feeding rates}}
& \multicolumn{4}{c}{\textbf{Co-occurrence rates}} \\
\cmidrule(lr){2-5} \cmidrule(lr){6-9}
& \textbf{Mean $\downarrow$
} & \textbf{Median $\downarrow$
} & \textbf{SD $\downarrow$
} & \textbf{r $\uparrow$}
& \textbf{Mean $\downarrow$
} & \textbf{Median $\downarrow$
} & \textbf{SD $\downarrow$
} & \textbf{r $\uparrow$} \\
\midrule
CORVID & \textbf{9.000} & \textbf{8.131} & 7.342 & \textbf{0.582} & \textbf{0.041} &\textbf{ 0.019} & 0.057 & 0.654 \\
MegaDesc. & 13.14 & 8.604 & 13.08 & 0.505 & 0.056 & 0.028 & 0.063 & 0.557 \\
Random & 9.349 & 9.287 & \textbf{5.667} & 0.437 & \textbf{0.041} & 0.028 & \textbf{0.046} & \textbf{0.799} \\
\midrule
\textbf{Human} & \textbf{1.88} & \textbf{0.80} & \textbf{3.47} & \textbf{0.910} & \textbf{0.028} & \textbf{0.007} & \textbf{0.053} & \textbf{0.913} \\
\bottomrule
\end{tabular}
\end{adjustbox}
\label{tab:appspec_upper}
\end{table}

\section{Discussion and Limitations}

Application driven ML is increasingly important, which raises recent discussions on how computer vision innovations can be bridged to real world applications~\cite{rolnick_position_2024,chan_towards_2025}. The CHIRP dataset is a task-diverse dataset consists of existing data collected from an ongoing long-term system. This ensures that the dataset is directly relevant for automated individual-level behavioral monitoring in the wild. To further the goal of bridging computer vision and the application domain, we also introduce the application-specific benchmarks, a collection of independent test videos, with novel metrics to evaluate the ability of computer vision algorithms to extract biologically relevant measures. This new mechanism acts as a independent "system test" for computer vision algorithms, as errors in different parts of a pipeline can propagate in unpredictable ways~\cite{chan_towards_2025,pantazis2024deep}. While these newly proposed metrics are not meant to replace traditional task-specific benchmarking, they acts as a bridge to allow biologists to directly evaluate whether computer vision solutions are sufficient for application, and directly apply architectures appropriately. 

Another feature of application-driven ML is the availability of domain-specific knowledge when designing computer vision solutions~\cite{rolnick_position_2024}. We incorporate this concept throughout the design and benchmarking of the CHIRP dataset, by providing color ring segmentation and metadata on the probable birds to constraint the re-id problem. This added layer of complexity encourages future computer vision solutions to make use of the constraints of the study system, as we demonstrate with the CORVID, which identifies individuals purely from detecting the presence of color rings, contrary to traditional re-id approaches.

The CORVID pipeline is more flexible than traditional approaches as it does not rely on a gallery to be matched with, and only relies on a list of possible color combinations. However, benchmarks on CHIRP showed that the accuracy of the framework depends on the constraint where only limited birds can be present in a video (\autoref{tab:Reid_bench}), making the method unscalable to larger bird populations. Future work can combine image based methods to compute similarities of appearances as demonstrated in other bird species \cite{ferreira2020deep,xiao2023multi}. This will be important for birds as many birds change appearance over time or seasons. 

Now, we discuss some limitations with the CHIRP dataset. 
Firstly, in contrary to the ChimpACT dataset \cite{ma_chimpact_2023}, annotations presented in the current dataset are done in different data subsets. 
This stems from the annotation strategy we employ, focusing our manual annotation efforts to diverse frames instead of video sequences like in ChimpACT~\cite{ma_chimpact_2023}.
To provide a solution to this problem, we provide model-annotated 2D keypoints and segmentation on the re-id and action recognition datasets.


Next, compared to other action recognition datasets, the number of annotations and behavioral classes provided in the dataset is limited, and is highly skewed towards feeding behavior (\autoref{fig:action_rec}). This is primarily a reflection of the distribution of behaviors that jays perform on the feeder, as they primarily spend their time feeding. The specific behaviors required depends on the final use case, however, quantifying feeding and individual presence/ co-presence is valuable for long-term behavioral monitoring, in relation to food acquisition and understanding the evolution of social behaviors \cite{meltzer_ecological_2025,griesser_fine-scale_2015,cunha2021you}. In addition, other behaviors like vigilance on the feeder (head held upright to identify predators \cite{griesser_nepotistic_2003,griesser_vigilance_2009}) can be reliably extracted from body postures if 2D keypoint estimates are accurate, and thus the types of behaviors that can be extracted using this dataset is not limited by the action recognition dataset.

Finally, the CHIRP dataset only includes data from a single species and study system, making it hard to evaluate whether developed algorithms can be generalized. However, CHIRP is a first of its kind dataset of a wild long-term bird population, which is limited by the years of effort that is required to establish and collect. We hope that the approach for preparing the CHIRP dataset, together with the novel application-specific benchmarking can act as a blueprint for future work, on how to design datasets carefully to encourage computer vision algorithms to be readily applied to the application domain. With well designed datasets and algorithms, computer vision can potentially revolutionize individual level behavior monitoring for the study of animal behavior, conservation and beyond.

\newpage

\section{Acknowledgments}
This work is funded by the Deutsche Forschungsgemeinschaft (DFG, German Research Foundation) under Germany’s Excellence Strategy—EXC 2117—422037984, DFG project 15990824, DFG Heisenberg Grant no. GR 4650/21 and DFG project Grant no. FP589/20. MHW is supported by an Oxford
Brookes University Emerging Leaders Research Fellowship. We thank Francesca Frisoni and Jyotsna Bellary for doing additional annotations for the application specific benchmark. We thank all the researchers and field workers who worked in the Luondua Boreal Field Station over the years for their contributions to the long-term dataset.



{
    \small
    \bibliographystyle{ieeenat_fullname}
    \bibliography{main}
}


\end{document}